# BIM-GPT: a Prompt-Based Virtual Assistant Framework for BIM Information Retrieval

Junwen Zheng[a, *], Martin Fischer[b]

[a] *Stanford University, Department of Civil & Environmental Engineering, Stanford, CA, United States*
[b] *Kumagai Professor of Engineering and Professor in Civil & Environmental Engineering, Stanford University, Stanford, CA, United States*

## Abstract

Efficient information retrieval (IR) from building information models (BIMs) poses significant challenges due to the necessity for deep BIM knowledge or extensive engineering efforts for automation. We introduce BIM-GPT, a prompt-based virtual assistant (VA) framework integrating BIM and generative pre-trained transformer (GPT) technologies to support NL-based IR. A prompt manager and dynamic template generate prompts for GPT models, enabling interpretation of users' NL queries, summarization of retrieved information, and answering BIM-related questions. In tests on a BIM IR dataset, our approach achieved 83.5% and 99.5% accuracy rates for classifying NL queries with no data and 2% data incorporated in prompts, respectively. Additionally, we validated BIM-GPT's functionality through a VA prototype for a hospital building. This research contributes to the development of effective and versatile VAs for BIM IR in the construction industry, significantly enhancing BIM accessibility and reducing engineering efforts and training data requirements for processing NL queries.

[*] Corresponding author.
E-mail address: junwenz@stanford.edu (J. Zheng), fischer@stanford.edu (M. Fischer).

## 1. Introduction

Building information models (BIMs), one of the most important digital representations of a built asset across its life cycle [1], have become increasingly popular as a tool in the construction industry [2]. BIMs can integrate multi-disciplinary data, such as architectural, structural, mechanical, electrical, and plumbing, to support construction and facility management activities [3]. The integrated BIMs contribute to cost and time savings [4], reduce risks [5], and improve efficiency of operation and maintenance [6].

However, as BIM is becoming more complex due to the aggregation of large amounts of data from design to construction and operation, it is more challenging to efficiently retrieve information [7], particularly for many non-tech savvy users in the industry. Existing information retrieval (IR) strategies require users to have extensive knowledge of BIM technology [8]. Practitioners must master not only the complicated user interface, but also data structure, terminology, and structured query language. The lack of BIM expertise has been identified as one of the most prohibiting barriers to realizing the benefits of rich, up-to-date information in construction and operation [9,10].

Virtual assistants (VAs) have been proposed as a potential solution to the challenges of BIM IR by offering a natural language (NL) interface [11–13]. However, the development of VAs requires extensive efforts, and their functionality is currently limited [14]. To interpret NL queries, existing VAs rely heavily on either traditional natural language processing (NLP) methods [11,12] or machine learning (ML) methods [13,15]. The traditional methods require substantial engineering to customize syntactic and semantic analysis for different BIMs. Although ML methods are more versatile than traditional NLP methods, they depend on large amounts of labeled data and computationally expensive model training. Moreover, these VAs can only respond to users' queries within predefined syntactic patterns, and cannot answer more general questions about building components and their properties in BIM that help practitioners retrieve information [16]. Unfortunately, the technical resources and training data required are generally not available in construction projects and for facility management, hindering the implementation of VAs in practice.

The emergence of generative pre-trained transformer (GPT) models, a type of large language model, presents new opportunities to improve IR. GPT models can significantly reduce the engineering and data requirements for prototyping VAs compared to traditional NLP and ML methods [17,18]. GPT models have been pre-trained on a large corpus of texts, such as GPT-3 [19] and InstructGPT [20], and have demonstrated remarkable in-context learning abilities with the help of a textual "prompt" [21]. For example, ChatGPT can perform many NLP tasks, such as text classification, information extraction, summarization, and question answering, by providing prompts that contain task instructions and a few demonstration examples as context [22]. Because these tasks are essential functions for efficient IR, the prompt-based approach is now being explored for VA development in retail [23] and healthcare [24].

This paper introduces BIM-GPT, a prompt-based VA framework that integrates BIM and GPT to enable efficient IR using NL. We developed a prompt library and a prompt manager, which serve as the core module of the framework to dynamically generate prompts that provide context for GPT models to interpret user queries, summarize retrieved results, and answer BIM-related questions. Leveraging the advantage of GPT's in-context ability, our approach not only supports NL-based user interactions to improve accessibility of BIMs, but also significantly reduces the required effort compared with existing approaches. Specifically, it eliminates the need for the customization of syntactic and semantic analysis in traditional methods as well as model training in ML methods. To validate BIM-GPT effectiveness and versatility, we evaluated the framework on the augmented BINLQ [15], a BIM IR dataset that we further annotated for improved coverage and granularity. The evaluation results demonstrate that BIM-GPT can accurately interpret user queries, with high accuracy achieved when no data is incorporated in prompts. Furthermore, we found that incorporating as little as 2% data in prompts can further improve the accuracy of the system. The evaluation results demonstrate that BIM-GPT can accurately interprets users' queries with high accuracy achieved when no data is incorporated in prompts. Furthermore, the results indicate that incorporating as little as 2% data in prompts can further improve the accuracy of the system. Additionally, we developed a VA prototype based on the framework for a hospital building and validated its functionality.

The paper is structured as follows: Section 2 reviews the related literature. Section 3 introduces the VA framework, illustrating its architecture and components. Section 4 evaluates the effectiveness and versatility of the framework based on a BIM IR dataset. Section 5 describes the implementation details for the VA prototype of a hospital building. Finally, Section 6 concludes the paper, summarizing key findings and outlining directions for future research.

1. Literature Review

BIM technology has revolutionized the construction industry in recent decades. From mere computer-aided design tools, BIMs have evolved into comprehensive systems that manage building information and project data throughout a building's life cycle [25]. As BIM adoption continues to grow in the industry [26], efficient IR becomes increasingly important, playing a critical role in enabling stakeholders to efficiently access and utilize the rich data available from BIMs [27].

Therefore, this study examined three areas of related literature to provide the context of BIM IR, identify gaps of existing retrieval strategies, and introduce new opportunities. Section 2.1 reviews the challenges of existing retrieval approaches for BIMs that require deep BIM knowledge. Section 2.2 explores the NL-based approaches for BIM IR, including traditional NLP methods and more recent ML methods. Section 2.3 introduces GPT models, particularly their in-context learning and NLP capabilities. Lastly, Section 2.4 summarizes the gaps that remain in existing NL-based approaches due to the need for extensive engineering and data requirements.

2.1 Challenges of BIM Information Retrieval

Retrieving information from BIMs involves accessing and extracting relevant data from complex digital models. BIMs capture a large amount of multi-disciplinary data throughout the design, construction, and operation of a construction project. This data is essential for project stakeholders' decision-making [28]. It consists of not only building elements' geometry, but also their semantics, such as the material, manufacturer, location, or system [29]. However, existing retrieval approaches, such as 3D interfaces and semantic search, require users to have deep BIM knowledge, including proficiency in graphical user interfaces (GUIs), understanding of data structures and their terminologies, and familiarity with formal query languages.

The 3D interface approach leverages intuitive GUIs of BIMs, enabling users to interact directly with building elements and access relevant information linked to them [30]. The approach is commonly used in Autodesk Revit [31] and other BIM software in practice and has also been applied in many BIM-integrated systems. For example, [32] supported users in visual knowledge management, while [33] facilitates operation and maintenance tasks. However, as BIMs grow larger and more complex, users may encounter difficulties in navigating and manipulating intricate 3D GUIs, hindering efficient IR.

The semantic search approach is also used for IR based on the object-oriented structure of BIMs [34]. This approach utilizes well-defined data schemas, such as the Industry Foundation Classes [35], and applies formal query languages to retrieve the desired information by filtering specific semantic properties or characteristics of building elements. For instance, [36] enables querying for building information of IFC-based BIMs with SPARQL, while [37] proposes a more intuitive visual programming language for general filtering of IFC-based BIMs. However, the semantic approach requires users to have a deep understanding of BIM data structures and programming concepts, which is generally not the case for non-tech savvy practitioners, especially in the construction and facility operation phrases of a project.

## 2.2 Natural Language-based Approaches

Over the last few decades, NLP technology has been applied to BIM IR, alleviating the knowledge and expertise requirements for practitioners. NL-based approaches enable users to retrieve information from BIMs using everyday expressions. Because of the massive information contained in BIMs and a large variety of NL, accurately interpreting users' queries is crucial for the effectiveness of such an NL-based interface.

Early attempts generally applied traditional NLP methods, such as syntactic analysis [38] and semantic analysis [39], for natural language understanding (NLU). Customized functions and algorithms were developed to identify a user's intent, extract key phrases and then map them to a predefined representation used for BIM IR represented in a structured format [14]. These were then applied to various use cases. [7,8] developed a process of keyword extraction and International Framework for Dictionary (IFD)-based mapping, including tokenization, tagging, parsing, classification and mapping steps, to convert an NL utterance into entities and relationships in IFC schemas for querying BIM data. [40] developed a string-matching method based on the Levenshtein distance and Burkhard-Keller tree structure to extract the spatial geometric information of a BIM from users' utterances for fire emergency response. [12] developed an NLU algorithm to classify content words and an information extraction algorithm to locate the queried structured IFC data from BIMs. Although these approaches could enable construction practitioners to retrieve information from BIMs using NL, they require extensive engineering efforts to customize NLP features and models.

To further improve the performance of NLU, previous research developed BIM-specific ontology models for semantic understanding. For example, [41] constructed an IFC IR ontology and used local context analysis to support a search engine for retrieving online BIM resources. [42] built a domain ontology for architecture, engineering, and construction (AEC) and a processing procedure for keyword extraction, expansion, and mapping to retrieve BIM objects. [11,43,44] implemented a BIM ontology model and applied NLP syntactic analysis through word segmentation, part-of-speech tagging, keyword matching extension and keyword mapping steps, to convert NL queries into SQL for retrieving and manipulating BIM data. Although ontology models can help further improve the accuracy of semantic analysis, additional engineering efforts are needed to build those models, which is typically not practical in construction.

Because traditional NLP-based methods require extensive effort to customize the syntactic and semantic analysis, a more versatile approach is needed to effectively interpret NL queries for BIM IR. In recent years, thanks to great advancements of ML in NLP, researchers have developed ML-based approaches that can learn NLP features and models from training data sets, reducing the need for engineering customized features and algorithms.

One notable example of ML methods is Bidirectional Encoder Representations from Transformers (BERT) [45]. BERT is a powerful pre-training language representation model that can be fine-tuned for specific NLP tasks, allowing it to learn from large amounts of text data and achieve state-of-the-art performance on a range of NLP benchmarks. [15] has applied a robustly optimized BERT pre-training approach (RoBERTa) fine-tuned on the building information-related natural language queries (BINLQ) dataset for classifying NL queries of BIMs into predefined IR categories. Moreover, technology companies have developed commercial products that utilize ML methods for NLP, such as Amazon Alexa, Google Assistant, and IBM Watson, which have also shown potential for BIM IR. [13] developed a proof-of-concept prototype, using Amazon Alexa, to retrieve information from BIMs. However, while these ML-based approaches can eliminate the need for customizing NLP functions and accurately recognize NL utterances, they require a large amount of training data, which is generally not available and too expensive to collect on typical construction projects [46].

In addition to NLU, natural language generation (NLG) has been applied to further process the data retrieved from BIMs into an NL-based representation. For example, [47] extracted semantic content from structured BIM data and then applied syntactic sentence structure, a grammar rule to generate NL sentences, helping practitioners easily understand and utilize the information. Recent research has integrated NLU and NLG to develop VA approaches that support NL-based IR. [12,48,49] developed the intelligent building information spoken dialogue system (iBISDS), a VA that provides IR for construction practitioners via spoken natural language queries. However, the development of NLG modules that require additional engineering efforts to build NL response templates is not practical and only covers limited syntactic patterns.

Although the VA approaches enable searching for BIM elements and their properties through NL, practitioners still need to learn general knowledge about BIM in order to better utilize the technology [16]. With the growth of the Internet, such knowledge is more accessible with the help of search engines. Semantic web technologies could facilitate retrieving the information from online sources and integrate them with BIM for energy analysis [50] and cost estimation [51]. In addition, to reduce search effort and enable NL queries, recent research has developed an intelligent question answering (QA) system for general BIM knowledge using a BERT-based ML method [16]. For example, the QA system can retrieve relevant documents from a dataset to answer questions, such as "is BIM a process or a model?" and "What is BIM?". Nevertheless, the supplemental functionality of QA also requires extensive engineering efforts in building the semantic web features or collecting data for ML methods.

## 2.3 New Opportunities: Generative Pre-trained Transformer Models

NLP has made significant progress in the last few years, largely due to the development of large language models (LLMs). LLMs are a class of advanced ML models designed to understand and generate human-like language by pre-training on massive amounts of text data [52,53]. One of the most prominent LLMs is the GPT series, including GPT-3 [19] with 175 billion parameters, which have emerged as powerful tools in NLP due to their in-context learning ability. Building upon GPT-3, its successor models such as InstructGPT [20] and ChatGPT [22] further enhance in-context learning by following human instructions more effectively.

These GPT models are capable of interpreting and generating NL with minimal additional training data [54]. Their in-context learning relies on "prompts" provided by users to guide the model's responses. This prompt-based approach allows GPT models to excel in zero-shot and few-shot learning scenarios [55]. They have achieved state-of-the-art performance in many NLP tasks, including accurately categorizing text into predefined classes with limited training data [19], extracting specific information using contextual prompts [56], generating concise and coherent summaries of lengthy text passages [57], and comprehending and generating accurate responses to user questions based on context [58]. Leveraging the NLP capabilities of GPT models, the prompt-based approach is now being explored for VA development. For example, [24] developed a GPT-based chatbot for triaging and patient note summarization for a healthcare use case. [23] reviews GPT-based chatbots applied for real-world customer service use cases. To date, however, there has been no published research on applying this approach for BIM IR in the construction sector.

## 2.4 Summary of Gaps in Existing Natural Language-based Approaches to BIM IR

The existing literature highlights the ongoing challenges of BIM IR, specifically the need for deep BIM knowledge, including proficiency in BIM GUIs, understanding of data structures, and familiarity with formal query languages. To address the accessibility issue of BIMs, NL-based approaches have been developed to allow practitioners to retrieve information using everyday expression. However, the current approaches summarized in Table 1 have several limitations.

To accurately understand NL queries, early approaches applied traditional syntactic and semantic analysis methods that required extensive engineering effort to customize functions for different BIMs. Recent ML-based approaches, more versatile than the traditional ones, were developed to learn the NLP functions and models from training data of NL queries. However, the construction industry generally lacks such data, which is labor-intensive and expensive to collect. Furthermore, all approaches use template-based methods to deliver retrieved information from BIM, which requires additional technical efforts and limits the scope to predefined templates. Lastly, none of them can answer general questions about building components and their properties in BIM, which would require additional effort to construct the knowledge base.

Table 1. Current NL-based Approaches for BIM IR with Limitations

| Approach | Natural language understanding (NLU) | Natural language generation (NLG) | Knowledge source for question answering (QA) | Limitations |
|---|---|---|---|---|
| AI-based voice assistant for BIM [13] | Machine learning: Amazon Alexa Skills | Template-based | BIM (RVT format) | A, C, D |
| Intelligent QA system for BIM & AIOT [16] | Machine learning: Fine-tuned BERT model | Not available (N/A) | BIM-AIOT QA corpus (text paragraphs) | A, D |
| QA system for BIM information extraction [12] | Traditional NLP: Syntactic and semantic analysis | Template-based | BIM (IFC format) | B, C, D |
| BIM automatic speech recognition [11] | Traditional NLP: syntactic and semantic analysis + ontology model | Not available (N/A) | BIM (RVT format) | B, D |
| Intelligent building information spoken dialogue system [48] | Traditional NLP: syntactic and semantic analysis | Template-based | BIM (RVT format) | B, C, D |
| A NL-based approach to data retrieval for cloud BIM [8] | Traditional NLP: syntactic and semantic analysis | Not available (N/A) | BIM (IFC format) | B, D |

A = requires a large amount of training data
B = requires extensive engineering effort
C = depends on predefined templates to deliver information
D = cannot answer general BIM-related questions

Although current NL-based approaches can alleviate the problem of practitioners in the construction industry often lacking BIM expertise, construction projects typically do not have the technical resources and training data. The emergence of GPT models offers new opportunities to address these limitations by leveraging the prompt-based techniques, reducing engineering requirements for traditional methods and data requirements for ML methods. Therefore, it is necessary to investigate the potential of integrating GPT with BIM to develop a more effective, versatile VA framework for IR, improving accessibility of BIMs without prohibitively increasing engineering efforts and requiring massive training data.

## 2. Research Methodology

This paper presents BIM-GPT, a prompt-based Virtual Assistant (VA) framework for BIM Information Retrieval (IR), designed to overcome limitations of existing approaches found in the literature. The BIM-GPT framework offers a NL interface, coupled with BIM visualization, to facilitate efficient IR and interaction with 3D building components for construction practitioners. To harness the NLP capabilities of GPT models, our framework is designed to dynamically generate prompts for various use cases and interpret NL queries accurately. Given that BIMs encompass massive, multidisciplinary building data, the framework must effectively manage this information to address user queries. Consequently, the BIM-GPT framework consists of three modules: a web-based User Interface (UI) module, an NLP module, and a Data Management (DM) module (as shown in Fig. 1).

The UI module enables users to engage with the VA by entering NL queries and receiving NL responses through chat boxes while visualizing the retrieved results in a cloud-based BIM. Text queries are relayed from the UI module to the NLP module, where responses are subsequently generated. To interpret the user's query, the NLP module dynamically produces prompts and acquires generated texts from an external GPT server. Within the DM module, a cloud-based database hosts the building information and provides results in response to structured queries from the NLP module.

The following subsections elaborate on the design of the NLP module, with its validation discussed in Section 4. Section 5 addressed the implementation of the UI and DM modules, as well as the validation of the entire framework.

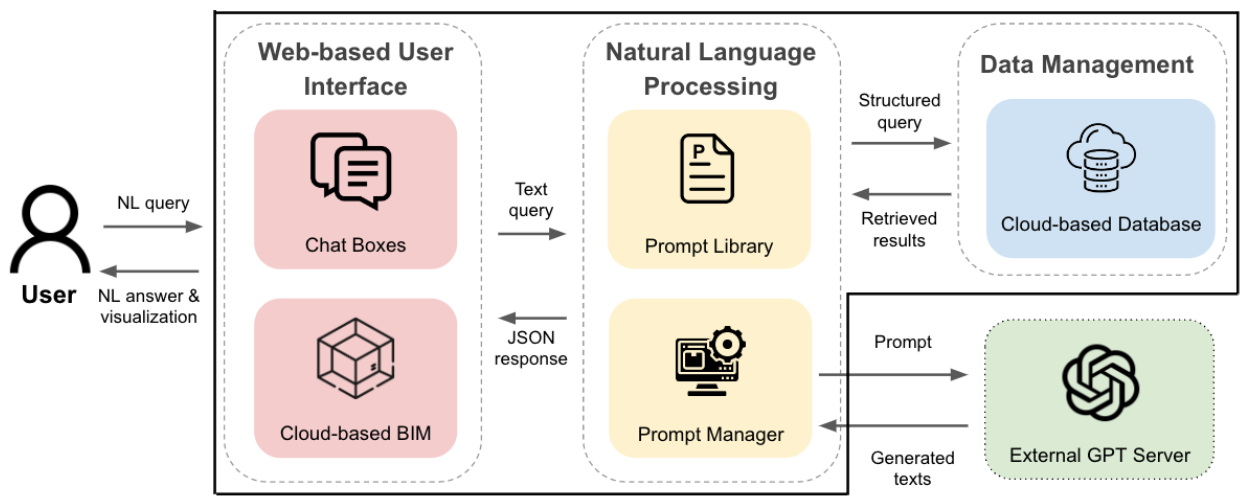

**Figure 1.** Overview of BIM-GPT Framework

### 3.1 Natural Language Processing Module

The NLP module serves as the core of the BIM-GPT framework, responsible for interpreting NL queries and generating NL answers based on retrieved results from BIMs. Our prompt-based approach leverages the NLP capabilities of GPT models to reduce engineering and data requirements. Trained on trillions of textual data to learn the patterns and structure of language, GPT models are designed to predict the most likely subsequent word when given a prompt as the starting point. The prompt is crucial in aligning the relevant data that the models have been trained on with the concepts in BIM IR, as depicted in Fig. 2, so that the models can generate relevant and coherent texts to meet users' needs.

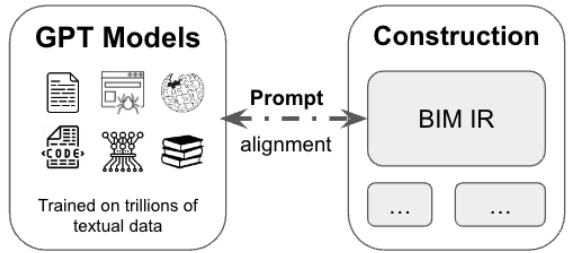

**Fig. 2.** Prompt: Aligning GPT Models with BIM IR

To support efficient IR and improve accessibility of BIMs, the NLP module must address three critical tasks: NLU, NLG, and QA. The NLU task entails classifying user intent and identifying the parameters as well as recognizing values of the queried building components, effectively converting NL queries into structured queries for IR from BIMs. The NLG task focuses on summarizing the retrieved results to deliver NL answers, while the QA task handles general questions related to building components and their properties within BIMs. Consequently, we designed a prompt library consisting of five types of prompts: Intent Prompt, Parameter Prompt, Value Prompt for NLU, Summary Prompt for NLG, and General Question Prompt for QA.

Furthermore, we developed a shared template for these five prompts. The most important feature of the template is that its components dynamically adapt based on the use case and user's query, which help GPT models effectively handle a wide variety of NL queries. Shown with the use cases in Fig. 3, the dynamic template consists of five components: System, Relevant Database Information, Task Instruction, Few-shot Examples, and User. This dynamic characteristic is essential for ensuring the framework's effectiveness and versatility across different NLP tasks.

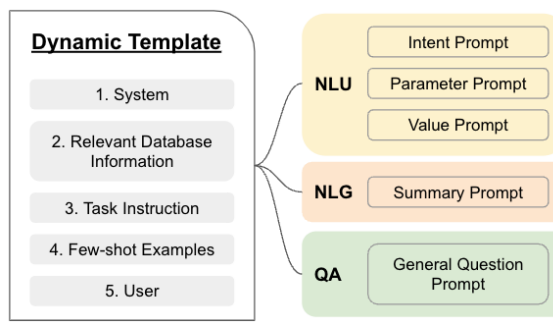

**Fig. 3.** Dynamic Prompt Template and Its Use Cases

To effectively manage the dynamic generation of prompts, we devised a prompt manager to control the process of converting NL queries into NL answers. The prompt manager is responsible for monitoring current states of NLP tasks, obtaining relevant information from the BIM database, updating prompts with retrieved results, and communicating with external GPT servers to generate texts based on the prompts. The following subsections elucidate the design of the dynamic prompt template in the prompt library and manager, supplemented by detailed examples.

### 3.1.1 Prompt Library

The effectiveness of the NLP module in the BIM-GPT framework relies heavily on the quality of prompts. To better align GPT models with BIM IR contexts, we designed the dynamic prompt template based on consideration of characteristics of GPT models. Because the prompt serves as the conditional probability for GPT models to predict the next word in a sequential order, the components of the template are arranged in a specific order, with the more pertinent content positioned closer to the end of the prompt. As such, the text towards the end of the prompt tends to carry greater weight than the text at the beginning, which helps GPT models generate relevant and coherent responses.

Following this sequence of relevance, we describe each component of the template and its role in aligning GPT models with BIM IR, along with an intent prompt example, as shown in Fig. 4. (1) The System component specifies the VA role of the GPT model, enabling it to understand the background of the user's query. For example, "You are a virtual assistant that helps users retrieve building information from the BIM database, and answer general questions about BIM technology, architecture, engineering, construction, and operation." (2) The Relevant Database Information component includes related data, such as schemas and records retrieved by the prompt manager from the BIM database that may provide hints for the GPT model. For example, "The BIM database consists of major building components of a hospital …" (3) The Task Instruction component describes the task and the output requirements in detail, which explicitly guide the GPT model to generate desired responses. For example, "Your first task is to classify the ONE intent … among …; Your second task is to identify ONE category … from the list …; This is a classification task; Answer as concisely as possible; The output should follow this template …" (4) The Few-shot Examples component provides examples of the user input and system output pairs, which demonstrate an output pattern for the GPT model. (5) The User component incorporates the user's query following the pattern, which serves as the starting point for the GPT model to generate texts. See Appendix A1 for additional examples of prompts.

**Intent Prompt Example**

**System**: You are a virtual assistant that helps users retrieve building information from the BIM database, and answer general questions about BIM technology, architecture, engineering, construction and operation.

---

**Relevant database information**:
The BIM database consists of all major building components of a hospital. For building components, the list of their attributes are: ['id', 'type', 'level', 'system_type', 'specification', 'room_name' …]

---

**Task instruction**:
Your first task is to classify the ONE intent of the user's query among [ask in GPT], [search in BIM], [count in BIM].
Your second task is to identify the ONE <category> of the building objects in the user query from the list: ['Air Handling Units', 'Chillers', 'Pumps', 'Transformers', …] or 'NA' if not found.
This is a classification task. Answer as concisely as possible.
The output should follow this template: A: [intent] of 'category'

---

**Few-shot examples**:
Q: What is BIM?
A: [ask in GPT] for 'NA'
Q: i'd like to find all the smoke detectors that serve for exhaust air 57 system.
A: [search in BIM] for 'Smoke Detectors'
Q: hi ! find me the air terminal 2253
A: [search in BIM] for 'Air Terminals'
……

---

**User**:
Q: Who is the manufacturer of pump 14569?

**Figure 4.** Intent Prompt Example of Dynamic Template

Even with the prompt template, it is still challenging for GPT models to directly parse an NL query into a structured query for BIM IR, because they have not been trained on such semantic parsing tasks nor seen the data structure of BIMs. To address this difficult NLU problem, we decompose it into a chain of subproblems, including text classification and information extraction and prediction, which can be handled by the models with appropriate prompting. Specifically, the GPT model first classifies the user intent and the category of building objects with the Intent Prompt, then identifies the parameters of the categorized building objects with the Parameter Prompt, and lastly recognizes the values of the identified parameters with the Value. In addition, the GPT model utilizes the Summary Prompt to summarize the retrieved results for NLG and the General Question Prompt for answering general questions about BIM technology, architecture, engineering, construction, and operation.

The dynamic nature of the prompt template is vital for the versatility and effectiveness of the framework. Because GPT models have a maximum number of words allowed in a prompt, supplying more pertinent information in prompts can assist models in interpreting NL queries more accurately and generating NL answers more coherently. Therefore, our template dynamically generates a prompt based on the use case and user query. The goal of the use case determines the general description of components for each type of prompt, while the user query further refines the exact details of relevant information to be incorporated into the prompts. Utilizing the same template can reduce engineering effort for developers and minimize learning costs for practitioners. The dynamically customized prompts can improve the performance of NLU, NLG, and QA tasks for BIM IR.

Table 2 presents the five use cases of prompts, with descriptions emphasizing how their components dynamically update according to the user query. Considering the large amount of building information stored in the database and the need for concise prompts, we devised a strategy to determine which pertinent information to include in the prompt components based on a user query. Firstly, the Task Instruction component offers all possible solutions of the task by employing known information. For example, the Intent Prompt includes the complete lists of predefined user intent and the list of all categories of building objects from the database. Secondly, the Relevant Database Information provides more detailed and granular information, such as database schemas and value records, compared to the specific information in the Task Instruction component. For instance, for each category (table) of building objects, the types and parameters represent a more detailed level of information. For each parameter of the categorized building objects, its distinct value records represent an even more specific level of detail. Lastly, the Few-shot Example includes relevant examples based on the identified category and parameter information. This dynamic design can enhance the performance of GPT models by aligning them more effectively with BIM IR contexts.

**Table 2.** Goals and Dynamic Nature of Prompts

| Use Case | Goal | Task Instruction | Relevant Database Information | Few-shot Example |
|---|---|---|---|---|
| Intent Prompt | Classify the intent and identify the queried category of building objects | • A list of all predefined user intent<br>• A list of all categories of building objects from the database | For each category:<br>• A list of its types<br>• A list of its parameters | Independent of user query |
| Parameter Prompt | Identify the filter and projection parameters | For the identified category:<br>• A list of its parameters | For each parameter of categorized building objects:<br>• A few example values | Examples that query the identified category of building objects |
| Value Prompt | Extract and identify the values of the filtered parameters | For the identified filter parameter of categorized building objects:<br>• A list of all its value records from the database | For the identified filter parameter of building objects:<br>• A few example records | Examples that query the identified filter parameter of the categorized building objects |
| Summary Prompt | Summarize the retrieved results from the BIM database | Independent of user query | For the query of identified category, parameters, values:<br>• A list of retrieved results from the BIM database | Independent of user query |
| General Question Prompt | Answer general questions about BIM technology | Independent of user query | Independent of user query | Independent of user query |

### 3.1.2 Prompt Manager

To dynamically generate prompts for each task, we have developed a PM to manage the entire process. Fig. 5 illustrates how the PM converts an NL query into an NL answer. At each step shown in yellow boxes, the PM builds the respective prompts, sends them, and receives generated responses from the GPT model through application programming interfaces (APIs). The PM uses the classification result of user intent to control whether to invoke either the BIM database search or the GPT general question answering. In addition, the PM makes API calls to retrieve relevant information from the BIM database (shown in the blue box), when constructing the prompts for parameter and value identification as well as summarization.

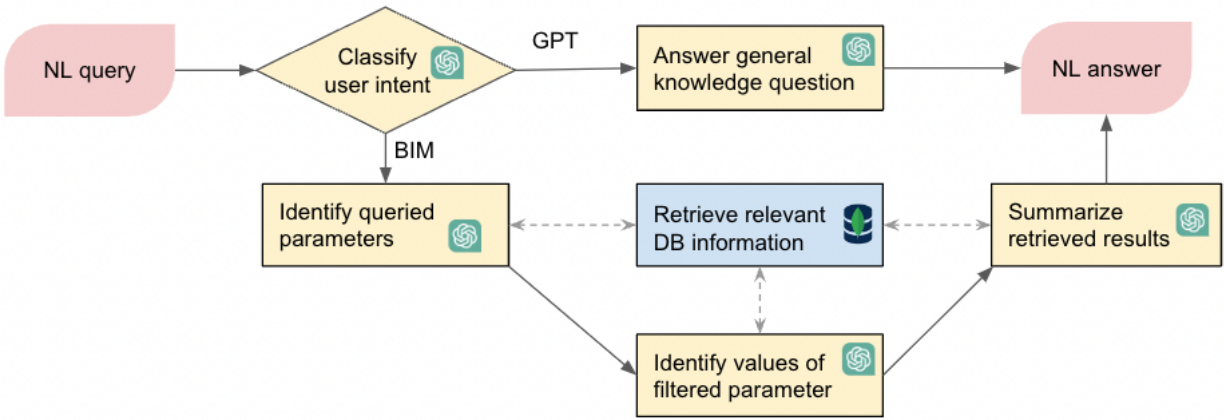

**Figure 5.** Prompt Manager: Converting NL Query into NL Answer

Fig. 6 shows a step-by-step example of how the PM interprets an NL query and generates a coherent NL answer, by incorporating pertinent information from the database into prompts and leveraging NLP capabilities of GPT models. For the query "Who is the manufacturer of pump 14569?", the process is listed below:

1. PM applies the Intent Prompt and identifies the intent as "[search in BIM]", and the category of the queried building objects as "Pumps";
2. PM retrieves all parameters of pumps and samples up to 10 values for each parameter from the database to build the Parameter Prompt, and identifies the filter parameter as "component_id" and the projection parameter as 'manufacturer';
3. PM retrieves a list of all component_id values for pumps to build the Value Prompt, and extracts the value from the user query as "14569" and predicted value as '14569' from the provided list of values;
4. PM retrieves the record of the queried pump from the database, constructs the Summary Prompt, and then generates the NL response: "The manufacturer of pump 14569 is PACO. It is located in room 06-470 on level 6 and is part of the hydronic return and power systems."

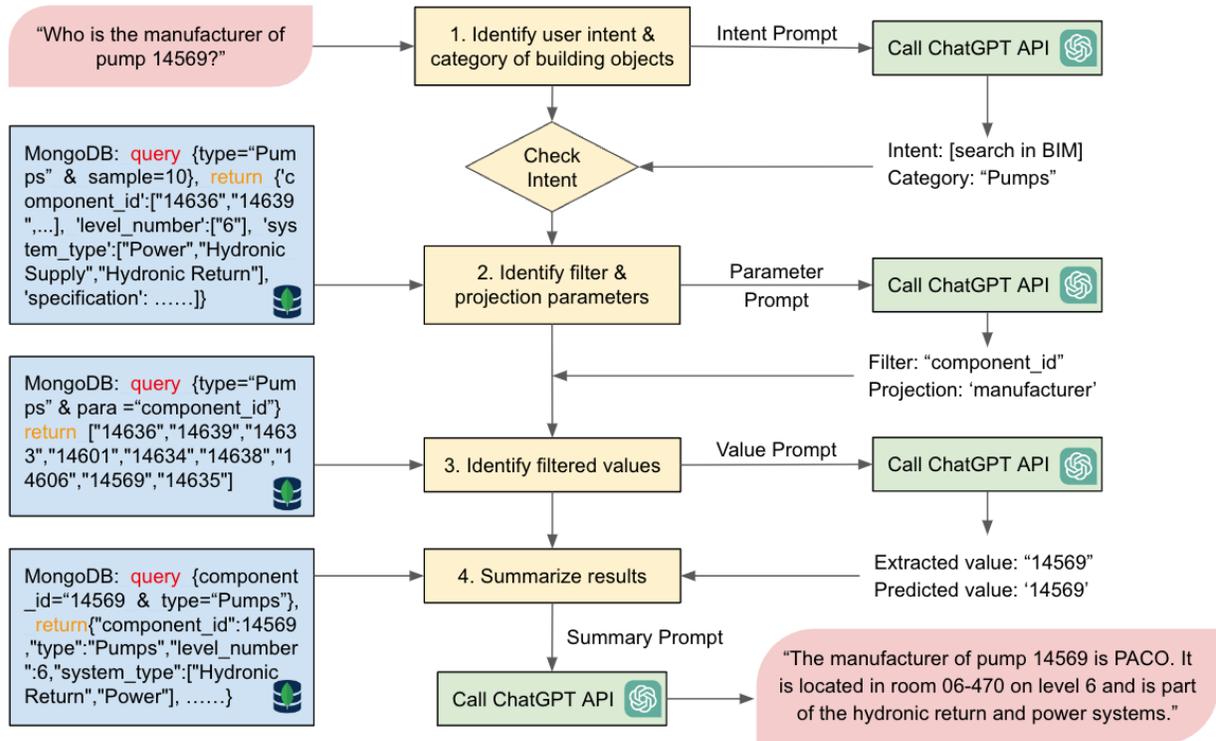

**Figure 6.** Example of How PM Processes NL Query

## 3. Validation of NLP for BIM-GPT

Accurately interpreting a user query is the prerequisite for VAs to effectively retrieve information from BIMs. This section reports on the experiments we conducted on a BIM IR dataset to validate the performance of NLP tasks for BIM-GPT. Our primary objective was to assess the accuracy of the framework, not only for classifying the intent and categories of building objects but also for identifying queried parameters and recognizing parameter values in NL queries. In addition, we tested the data efficiency of BIM-GPT under the zero-shot and few-shot scenarios, where no data was incorporated in prompts and 2% NL queries from the dataset were incorporated in prompts, respectively. Compared with the existing state-of-the-art approach that used 80% of the same dataset for training ML models and achieved 99.8% accuracy [15], BIM-GPT achieved comparably high accuracy of 99.5% in the few-shot scenario. Furthermore, we performed an ablation study for the dynamic template to understand the impact of prompt components on system performance. Besides NLU, the other NLP capabilities of BIM-GPT, such as NLG and QA, will be examined in Section 5.

### 4.1 Experiments

Our experiments utilized the BINLQ dataset [15], which we further annotated for improved coverage and granularity. The existing dataset includes 2,065 NL queries for two architectural BIMs, classified by the user intent and building object category. It includes 11 text classification (TC) labels related to the basic model information, attributes and quantity information of building objects such as windows, doors, rooms, units, and building storeys, as well as out-of-domain queries from the SQuAD dataset [59]. However, to retrieve exact information from BIMs, VAs must identify the parameters and their corresponding values in NL queries, in addition to the intent and category of building objects. While intent classification is important, the subsequent parameter identification and value recognition are crucial in obtaining specific information from NL queries to construct structured queries. These tasks are more difficult as the number of possible solutions increases from intent and categories to parameters and values.

To evaluate the effectiveness and versatility of our framework, we augmented the existing dataset by manually annotating the parameters and values for NL queries. We provided both the filter parameter (filter_para) and projection parameter (proj_para), along with the value extracted from the user query (extr_value) and the predicted value (pred_value) matching the database record. Since most databases only contain one distinct record for each value, the predicted value can be considered as the normalized form for the extracted value that may be its synonym or plural form. For example, for the query "What is the elevation of the second floor?"; the filter_para is "storey_id"; the proj_para is "elevation"; the extr_value is "second floor"; and the pred_value is "2". Our additional annotation at the parameter level increases the granularity of the dataset, extending beyond existing TC labels at the intent and category (table) level. Table 3 displays examples of NL queries with additional annotations in columns 4 to 7.

**Table 3.** Examples of Additional Annotations for NL Queries in BINLQ [15]

| NL Query | Intent | | Parameter | | Value | |
|---|---|---|---|---|---|---|
| | TC Label | Category (table) | Projection Parameter | Filter Parameter | Extracted Values | Predicted Values |
| How wide is the door in the faculty office 0331? | ATT-DOOR | door | width | room | "faculty office 0331" | "faculty office 0331" |
| What is the elevation of level 2? | ATT-STOREY | storey | elevation | storey_id | "level 2" | "2" |
| Tell me the windows' object type in faculty office 0332. | ATT-WINDOW | window | object_type | room | "faculty office 0332" | "faculty office 0332" |
| What is room 0201's long name? | ATT-ROOM | room | long_name | room_id | "0201" | "0201" |
| What is the length unit's name? | ATT-UNIT | unit | name | unit_type | "length" | "length" |
| How many windows are in medium classroom 0106? | QTY-WINDOW | window | quantity | room | "medium classroom 0106" | "medium classroom 0106" |
| What is the number of doors is in the medium classroom 0306? | QTY-DOOR | door | quantity | room | "medium classroom 0306" | "medium classroom 0306" |
| What are the quantities of building storeys in the Rinker model? | QTY-STOREY | storey | quantity | bim_file | "Rinker" | "rinker model" |
| What are the quantities of medium classrooms in the model? | QTY-ROOM | room | quantity | room_type | "medium classrooms" | "medium classroom" |

With the augmented dataset, our framework was implemented using Python (version 3.10.8) for the experiments. We used the ChatGPT model (gpt-3.5-turbo-0301) and set its temperature parameter to 0 to minimize the randomness of text generation. The Python Data Analysis (Pandas) library was used to handle this dataset. For each building object category in the dataset, its schemas and distinct records of those schemas were retrieved and organized in a Python object, along with NL queries and labels stored in a Pandas dataframe. We applied the dynamic template to the intent, parameter and value prompts, which are updated according to the user query. Because we decompose the problem of semantically parsing NL queries into a chain of subproblems, only the NL queries accurately interpreted from the intent classification and the parameter identification were proceeded for the parameter identification and the value recognition, respectively.

Given a specific dataset, its categories (tables), schemas, and distinct values are known information that can be included in the Relevant Database Information and Task Instruction components of prompts. However, the NL queries and their labels require labor-intensive data collection, which is generally not available for construction projects. To validate the data efficiency of our approach, we established the zero-shot and few-shot scenarios for incorporating NL queries and their ground truth labels in the Few-shot Example components of prompts. In the zero-shot scenario, we included one or two hypothetical examples, such as NL queries not included in the dataset and the queries failed in the preceding tasks, to demonstrate the output pattern for the GPT model, which did not affect the task accuracy. Table 4 Illustrates the example preparation for the few-shot scenario. For the Intent Prompt, the 2% data sampling was independent of the user query, while for the Parameter Prompt and Value Prompt, the sampling was based on the identified category and the identified parameter, respectively. It is important to note that the few-shot learning was much more data efficient than most ML-based methods. For example, [15] leveraged transfer learning techniques to reduce data requirements and used 80% of the same dataset to train ML models for classifying TC labels. See Appendix A1 for examples of prompts.

**Table 4**. Example Preparation for Few-shot Scenarios

| Use Case | Sampling Methods |
|---|---|
| Intent Prompt | Randomly sampled 2% NL queries for each TC label in the dataset (fixed set for ALL NL queries) |
| Parameter Prompt | Randomly sampled 2% NL queries for each parameter of the categorized building objects (fixed set of NL queries for a specific object category) |
| Value Prompt | Randomly sampled 2% NL queries for each distinct value record for the identified parameter (fixed set of NL queries for a specific parameter of categorized objects) |

To investigate the impact of different prompt components on task performance, we conducted an ablation study of the dynamic template. Specifically, we systematically removed one component at a time, including the System, Relevant Database, and Task Instruction components, and evaluated the accuracy of the prompts in intent classification, parameter identification, and value recognition. Additionally, the ablation study was performed under both zero-shot and few-shot scenarios to further comprehend their influence with respect to the data availability in the Few-shot Example component. This study not only benefits VA developers but also construction practitioners, as both parties can participate in editing and refining the NL-based prompts. The results show that the Task Instruction component was the most critical component and collecting labeled data for the Few-shot Examples was useful for accurately interpreting NL queries for BIM IR. In the following section, we present the results and discuss their implications.

## 4.2 Results

Table 5. presents the results of our experiments on the framework. In terms of the Intent Prompt, BIM-GPT achieved accuracy rates of 83.5% and 99.5% for classifying TC labels under the zero-shot and few-shot scenarios, respectively. For building object category classification, a subclass of TC labels, the zero-shot accuracy increased to 98.6%, while the few-shot accuracy remained almost unchanged. For identifying filter and projection parameters, the zero-shot accuracy rates were 92.1% and 90.1%, respectively, while the few-shot accuracy rates were 97.8% and 98.2%, respectively.

Regarding the Value Prompt, BIM-GPT achieved accuracy rates of 78.3% and 88.9 % for predicting the values of filtered parameters under the zero-shot and few-shot scenarios, respectively. The framework also achieved accuracy rates of 62.0% and 81.7% for extracting the values from NL queries under the zero-shot and few-shot scenarios, respectively. Notably, the accuracy of "Pred. / Extr. values," which refers to cases where the GPT model correctly predicted or extracted the value, was higher than each individual accuracy rate. This suggests that the extracted values have the potential to further improve the value prediction. Overall, our prompt-based approach performs well in NLU tasks, achieving high accuracy rates.

**Table 5**. Accuracy Rate of NLU Tasks

| Task Accuracy | Zero-shot | Few-shot |
|---|---|---|
| TC label | 83.5% | **99.5%** |
| Object category | 98.6% | 99.5% |
| Filter parameter | 92.1% | **97.8%** |
| Projection parameters | 90.1% | 98.2% |
| Predicted values | 78.3% | **88.9%** |
| Extracted values | 62.0% | 81.7% |
| Pred. / Extr. values | 86.7% | 95.1% |

The confusion matrices for TC labels are presented in Fig. 7, which illustrates the classification table between true and predicted labels. As shown on the left, the zero-shot prompt tended to misclassify the attribute and quantity label while accurately predicting the category of building objects. On the right, the clear diagonal concentration indicates that the few-shot prompt correctly classified most TC labels. Error analysis of the results revealed a common mistake where the model misclassified NL queries such as "how high is double glass K?" and "What is the width of rinker double d" as the window category instead of the door category. However, these ambiguous NL queries are also confusing for humans.

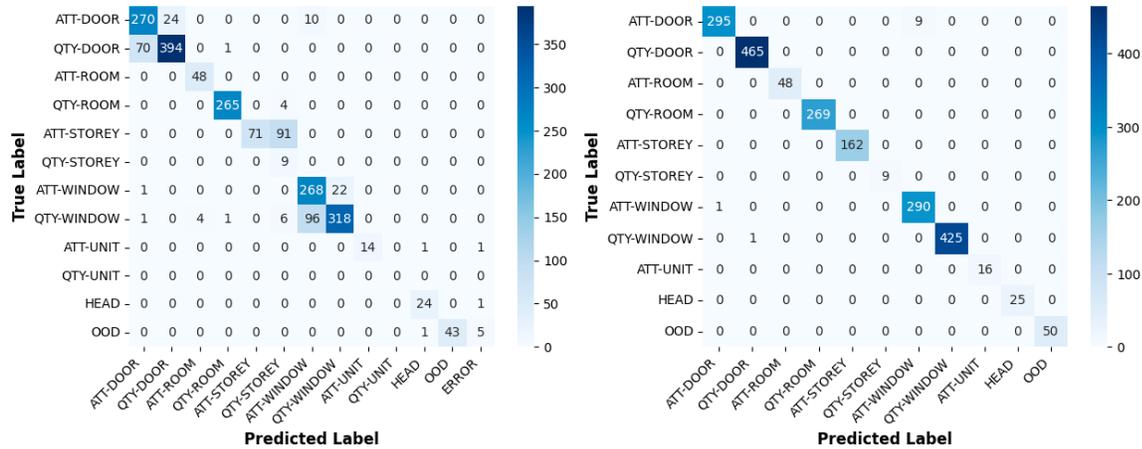

**Figure 7.** Confusion Matrices for TC Labels: Zero-shot (left) & Few-shot (right)

To provide insights into BIM-GPT's performance in identifying and predicting parameters and values, we present the categorized results in Fig. 8. Predicting the exact values that match distinct database records is a more difficult task, and this is reflected in the decreased accuracy rates for filtered values, which were 15% and 9% lower than the rates for filtered parameters under the zero-shot and few-shot scenarios, respectively. As shown on the left of Fig. 8, the accuracy rates for filtered parameters were on average over 90% for most building object categories. For the zero-shot scenario, the unit category has the lowest accuracy rate of 87.5%, but it only contained a small number of NL queries. The door category, with 89.7% accuracy, contributes most to the misclassification, followed by the room category, with 89.3% accuracy. The average few-shot accuracy was 6% higher than the zero-shot one, indicating that providing labeled data was effective to improve the performance of parameter identification.

On the right of Fig. 8, the accuracy rates for filtered values were generally above 80% for most categories, with the exception of the door category. In the zero-shot scenario, the accuracy rates of value recognition for the door and window categories were the lowest, at 72.2% and 80.4%, respectively. However, the average few-shot accuracy for value recognition was 14% higher in the few-shot scenario compared to the zero-shot scenario. This indicates that incorporating a small amount of NL queries with annotations was useful for the GPT model. Error analysis did not reveal any typical patterns, and one of the major challenges the model faced was the large number of distinct records in the database for the "room" and "id" parameters. For instance, the "id" and "room" parameters had over 50 and 150 different values, respectively, making it difficult for the GPT model to accurately identify the filtered value.

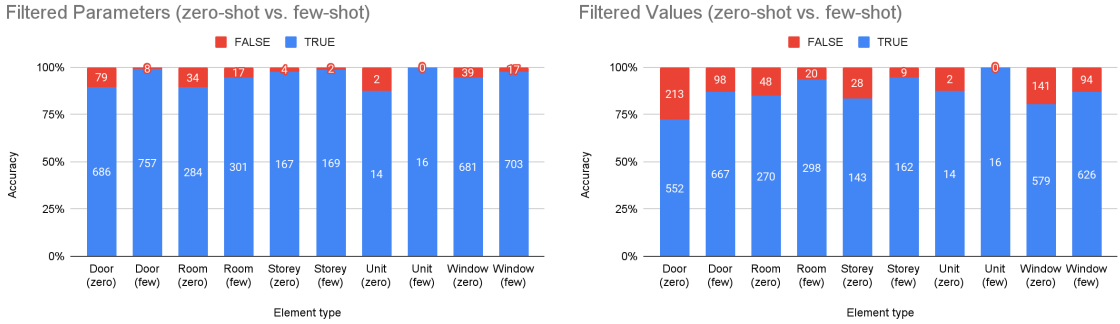

**Figure 8.** Categorized Filter Parameters (left) & Value Prediction (right)

Table 6 presents the results of our ablation study, in which we evaluated the impact of the System (SYS), Relevant Database Information (DB), and Task Instruction (TASK) components of the dynamic prompt template on the performance of the NLU tasks. We measured the accuracy of classifying TC labels, object categories, identifying filter and projection parameters, and extracting and predicting filtered values for both zero-shot and few-shot scenarios, while removing one component at a time. As expected, the effect of removed components increased following their sequences in the template for the zero-shot scenario. Specifically, removing the TASK component significantly degraded the accuracy rates for all tasks (shown in red). For most tasks in the few-shot scenario, the DB and SYS components did not influence the accuracy rates as much as the TASK one (shown in bold text). However, removing the TASK component did not affect the accuracy for value prediction and extraction (shown in blue).

**Table 6.** Ablation Study Results

| Prompt Composition | TC label (zero) | Category (zero) | TC label (few) | Category (few) | Filter_para (zero) | Proj_para (zero) | Filter_para (few) | Proj_para (few) | Pred_value (zero) | Extr_value (zero) | Pred_value (few) | Extr_value (few) |
|---|---|---|---|---|---|---|---|---|---|---|---|---|
| SYS + DB + TASK + FEW | 83.5% | 98.6% | 99.5% | 99.5% | 92.1% | 90.1% | 97.8% | 98.2% | 78.3% | 62.0% | 88.9% | 81.7% |
| DB + TASK + FEW | 83.9% | 98.5% | 99.4% | 99.5% | 90.9% | 90.0% | 97.3% | 98.2% | 78.9% | 62.1% | 88.6% | 81.8% |
| TASK + FEW | 75.2% | 83.8% | 99.2% | 99.0% | 79.2% | 87.0% | 95.9% | 98.1% | 81.2% | 65.0% | 88.7% | 81.8% |
| FEW | 0.0% | 0.0% | 71.4% | 72.6% | 25.2% | 59.1% | 83.1% | 88.1% | 23.4% | 31.6% | 89.2% | 81.8% |

Figure 9 presents the results of the ablation study for the Intent Prompt. Removing the SYS components had no effect on task accuracy rates for either scenario, as the SYS only describes background and role information for VAs. However, such descriptions might be useful to stop the GPT model from generating irrelevant texts. For example, in the query "Is there a louver 37 on 01 fl 01 t.o. Slab?", the intent prompt without the SYS component generated irrelevant user queries and answers following the correct classification. Additionally, removing the DB component resulted in a 9% and 15% reduction in accuracy for the TC label and object category tasks, respectively, in the zero-shot scenario, while it had no impact in the few-shot scenario. Although the type and attribute information of categorized building objects in the DB could provide some context for the GPT model, this information was less useful compared to more specific few-shot examples. Without the TASK component, the zero-shot prompt did not generate any meaningful output, as the accuracy rates dropped to 0%. Nevertheless, by providing only 2% of the data as few-shot examples, the accuracy rates only dropped to about 70%. This indicates that collecting NL queries, even just a small number of them, is useful in classifying the user intent and object category.

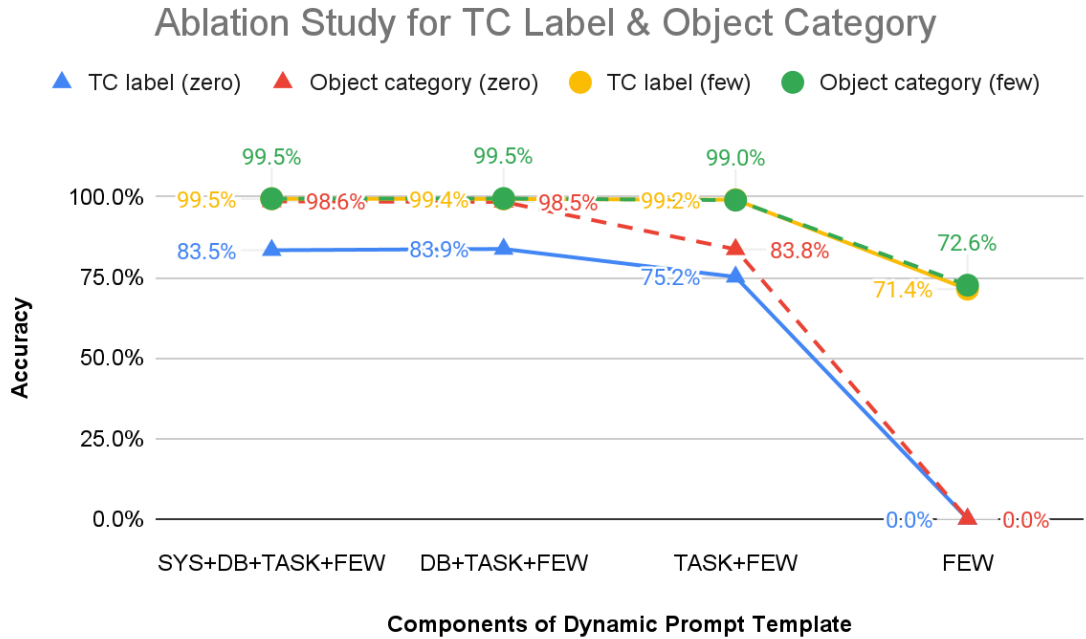

**Figure 9.** Ablation Study Results for Intent Prompt

Figure 10 illustrates the ablation study results for the Parameter Prompt. In the zero-shot scenario, removing the DB component did not affect the accuracy of identifying projection parameters, as our annotation only included the projection parameters but not their value records. However, removing the DB component decreased the accuracy of identifying filter parameters by 11%. Furthermore, removing the TASK component greatly reduced the accuracy rates for identifying projection and filter parameters to 59.1% and 25.2%, respectively. Despite this, because we added leading texts like "proj_para for '' and "filter_para for" in front of NL queries in the prompts, the GPT model may have had a rough idea of the task even without the TASK component. In the few-shot scenario, only the TASK component affected the accuracy of parameter identification, with around a 10% reduction. These results indicate that the TASK is the most critical components for parameter identification. Thus, when construction practitioners, with rich domain expertise participate in the prompt development process, they should ensure the proper design of the TASK component. This can help achieve an acceptable accuracy rate, which is 80% for our case, and the collection of few-shot data can further improve accuracy, closing the remaining 20% gap.

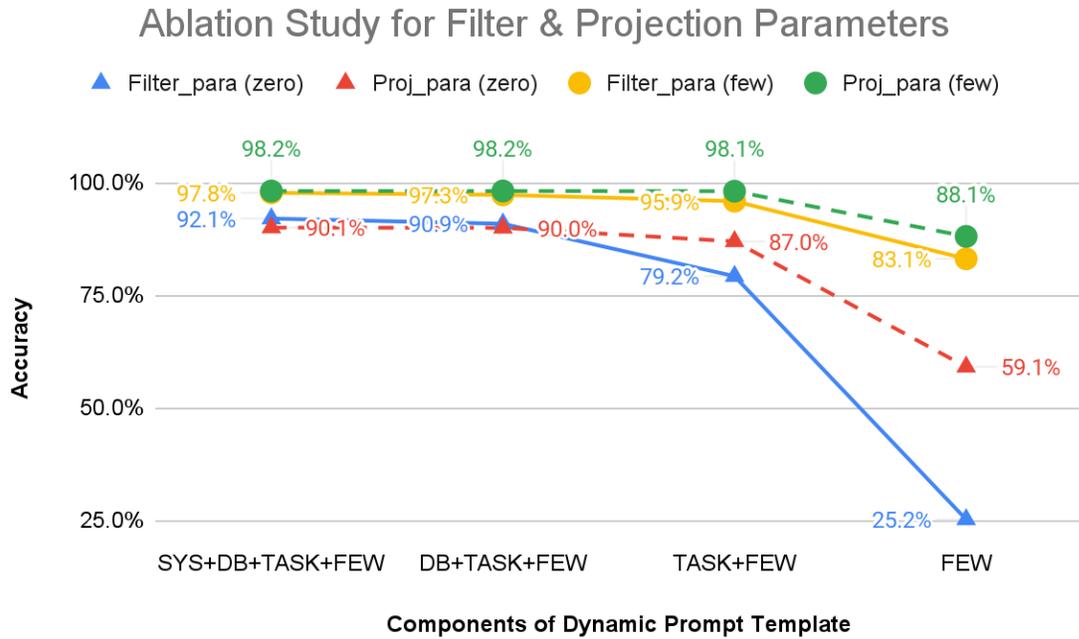

**Figure 10.** Ablation Study Results for Parameter Prompt

Figure 11 displays the ablation study results for the Value Prompt. In the zero-shot scenario, removing the SYS component did not affect the accuracy of filtered value prediction and extraction, while removing the DB component slightly increased the accuracy. This suggests that including the values of the same parameter for other categorized objects in the DB might introduce noise to the GPT model. For the few-shot scenario, none of the SYS, DB, and TASK components affected the accuracy of filtered value prediction and extraction. This may be because the few-shot examples already provided adequate context and information for such specific tasks. Additionally, the accuracy rates for prediction were higher than those for extraction, which may be due to the fuzzy definition of the ground truth values. For example, consider the query "What is the object type of the door in room 0220?" The annotation of the extracted value was "0202", while the GPT model extracted "room 0220", resulting in an incorrect classification.

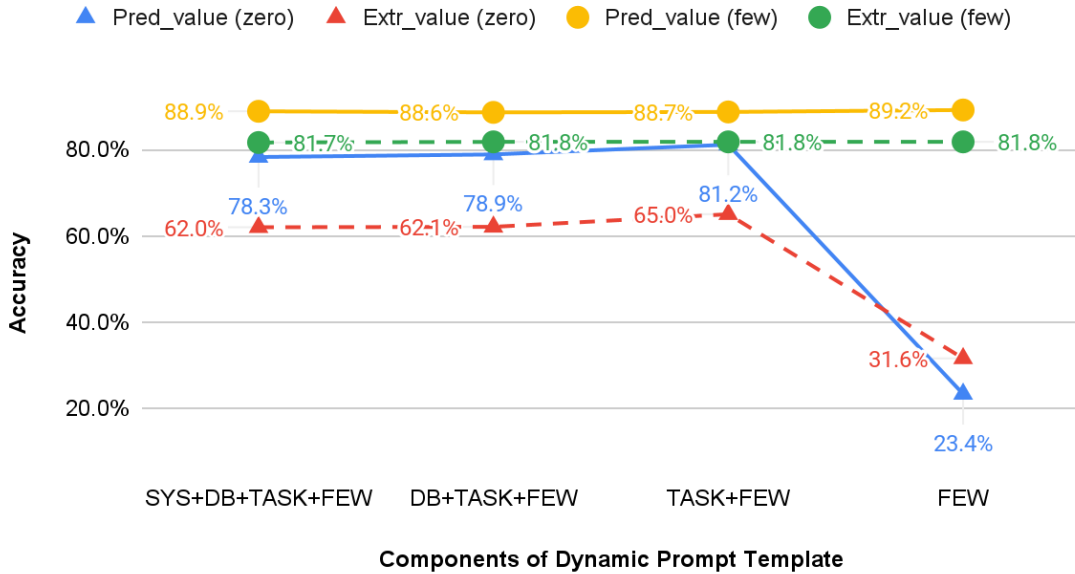

**Figure 11.** Ablation Study Results for Value Prompt

### 4.3 Discussion

The accurate interpretation of user queries is critical for NL-based approaches for BIM IR. Our experiment results demonstrate the effectiveness and versatility of the BIM-GPT framework for a range of NLP tasks. By aligning the GPT model with BIM IR context, our prompt-based approach successfully classified the intent and building object category, identified filter and projection parameters, and extracted and predicted parameter values from NL queries. Our dynamic template was successfully employed to generate the Intent, Parameter, and Value Prompts, and achieved high accuracy rates, even in the zero-shot scenario. Additionally, we further enhanced the accuracy by incorporating supplementary examples in the few-shot scenario.

The experimental results also substantiated the data efficiency of our approach in comprehending NL queries. In comparison to the existing state-of-the-art method [15], which necessitated 80% of the same dataset for training purposes and attained 99.76% accuracy for the remaining 20% data, BIM-GPT achieved accuracy rates of 83.5% and 99.5% with no data and a mere 2% of data incorporated in prompts, respectively, when tested on the complete dataset. This indicates that BIM-GPT can more efficiently utilize the collected data for testing, ensuring its robust performance in covering a wide variety of NL queries.

The results of the ablation offer insights into the influence of the components of the dynamic prompts on system performance. We discovered that the Task Instruction component was most critical, providing the necessary context for the GPT model to generate accurate responses. The Relevant Database Information component was particularly beneficial for identifying object categories and parameters, especially in zero-shot scenarios. Although the System component did not considerably impact the system's performance, it may aid in preventing irrelevant text generation of the GPT model. This assessment of prompt components can not only help construction practitioners prioritize those components to improve task performance, but also guide the future development of more effective prompts.

In summary, our prompt-based approach exhibited excellent performance for text classification tasks, such as intent classification and parameter identification, and reasonably good performance for information extraction and prediction tasks, such as value recognition. By leveraging the NLP capabilities of the GPT model, BIM-GPT achieved up to 98% accuracy for classifying texts from a limited number of predefined labels (e.g., intent, categories, parameters). However, attaining the same level of accuracy for predicting specific values from a large number of possible solutions remained challenging for BIM-GPT. Furthermore, although our prompt-based approach was much more data efficient than existing approaches, incorporating labeled data in prompts still determines how accurately NLP tasks are performed, especially difficult ones.

### 4.4 Limitations

Although our framework's results demonstrate promise, several limitations need to be acknowledged. Firstly, the evaluation of the framework was conducted on a single BIM IR dataset. This augmented dataset covered only NL queries for attribute and quantity information of a few categorized building objects, such as windows doors, rooms, units and stories, and a limited range of filter parameters and values. Geometric information about building objects, such as width, length, and height, which are useful in construction projects, was not included. Future research should collect more NL queries for different BIMs to assess the framework's generalizability.

Moreover, the prompt-based approach is still a proof-of-concept method based on the recently released GPT model (i.e., ChatGPT). It remains challenging for the approach to perform accurately on difficult NLU tasks that involve numerous labels, particularly classification and prediction. Future work should improve the design of prompts to better align GPT models with the BIM IR context. Because the sequence and composition of few-shot examples and relevant information incorporated in prompts can influence GPT model performance, future research should also explore more effective sampling methods tailored to user queries. In addition, future studies should investigate how different GPT models may affect the performance of the BIM-GPT framework.

## 4. Validation of BIM-GPT Framework

In the previous section, we validated the effectiveness and data efficiency of the BIM-GPT framework in interpreting NL queries for BIM IR, demonstrating its potential for practical applications. Building upon these findings, Section 5 will present the validation of the complete BIM-GPT framework through the implementation of a VA prototype for a hospital building, showcasing its functionality and evaluating the integration of the DM, NLP, and UI modules.

## 5.1 Implementation

Based on the BIM-GPT framework, we developed a VA prototype for a sample hospital building. The hospital BIM comprises over 42,000 building objects, including 675 facility assets with rich and structured properties. The implementation process for BIM-GPT, illustrated in Fig. 12, involved three main steps: data preprocessing for the DM module; building and testing the prompt library and manager for the NLP module; and integrating the web-based interface for the UI module.

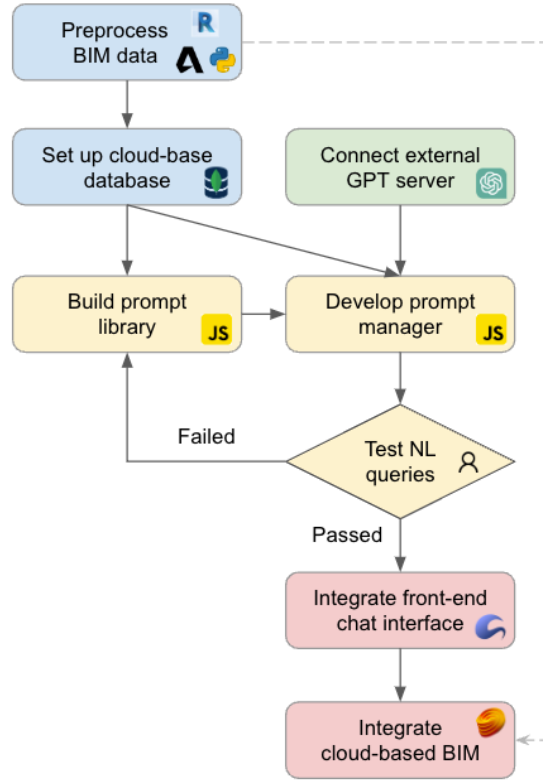

**Figure 12.** Implementation Process for BIM-GPT

### 5.1.1 Data Management Module

The development of the DM module involved a series of preprocessing steps to prepare for high quality data for BIM IR, as shown in Fig. 13. Firstly, the hospital BIM consisted of the architectural, structural, and mechanical models in the format of Autodesk Revit. Our first step was to use the Model Derivative API of the Autodesk Platform Service to extract building objects and their properties, and to translate the Revit model into SVF2 format that can be rendered in a browser. Next, we cleaned the extracted building information using Python to ensure data quality. The data attributes of the building objects were categorized into five groups: basic information (i.e., component_id, component_type, is_asset); location information (i.e., level_number, room_type, room_name); building system information (i.e., system_type, system_name); equipment information (i.e., manufacturer, model_name, specification); and OmniClass information (i.e., title, number). Finally, we imported the cleaned data into the cloud-based MongoDB database. The data was stored as documents in BSON format, which allows for flexible and efficient querying. Additionally, the translated model was used for the cloud-based BIM with Autodesk Forge API, enabling users to access the BIM through a web browser. Overall, the DM module effectively extracted and cleaned the necessary data from the hospital BIM, making it readily accessible for the subsequent NLP module.

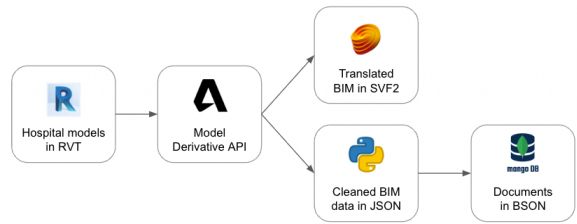

**Figure 13.** Data Preprocessing

### 5.1.2 Natural Language Processing Module

For the NLP module, we implemented the prompt-based approach using JavaScript and Node.js (v18.12.1) along with Npm (8.19.2). We built upon the methodologies outlined in Sections 3.1 and 3.2 to develop a comprehensive prompt library and a prompt manager that connected with the ChatGPT model (gpt-3.5-turbo) via the OpenAI API. To evaluate the NLP module, we tested it with a set of 100 NL queries. While Section 4 elaborates on the implementation of prompts for identifying user intent, parameters, and values, this section focuses on prompts for summarization and general question answering, and the NL-based prompt development process.

For the Summary Prompt, we incorporated the retrieved results from the BIM database and explicitly defined the task to answer user's queries The Task Instruction component stated: "Your task is to provide a summarized response of the retrieved results from BIM to answer the user's query. Your summary must be based on the retrieved results. You can use the first 1-2 as examples, where many records are retrieved. You may provide the id and type, location, and building system of the retrieved building components, which are generally useful for users." This task description effectively guided the GPT model in summarizing the retrieved results for various situations, thereby delivering the desired information to users.

For the General Question Prompt, we delineated the scope of general questions that the VA would address and utilized the GPT model as an external knowledge base. Specifically, the Task Instruction component stated: "The user may ask questions about the functions of building components and the definition of the components' properties in the BIM database; users may also search for general knowledge about buildings, BIM technology, architecture, engineering, construction, and facility management. For other questions, your answer should be 'Sorry, I do not know. Your question is out of my scope.'" This scope definition minimized the generation of irrelevant text by the GPT model, preventing potential user confusion about its functionality.

The final aspect of the prompt development involves testing and refinement of the prompts to ensure continuous improvement. BIM-GPT offers a significant advantage over existing approaches because it allows for easy improvement of the system's NLP capabilities by refining the prompts in the library, rather than necessitating extensive revisions for traditional NLP methods, such as syntactic and semantic analysis, and the need for retraining ML models. For such an NL-based development, even practitioners without programming backgrounds can participate in improving the prompts. Specifically, we incorporated failed NL queries into the Few-shot Example component of prompts to enhance the accuracy of NLU. The prototype was tested with 90 NL queries for the BIM database search, which we collected and manually annotated, and 10 general questions about BIM technology sourced from an existing BIM QA dataset [16]. Examples of general BIM questions included "Is the BIM a process or a model?", "What is the advantage of BIM?", and "How to define BIM?". Although we initially operated under a zero-shot scenario, the performance of the VA prototype showed steady improvement throughout the testing process.

### 5.1.3 User Interface Module

We developed a web-based UI for the VA prototype to support efficient BIM IR. To enable NL interaction, we integrated with the open-source Genie Server [60] developed by Stanford Open Virtual Assistant Lab (OVAL) as the front-end chat interface. The Genie Server was further integrated with Autodesk Forge, which provided a library for visualizing the cloud-based BIM. To enable seamless communication between the front-end interface and the prompt manager, we used web sockets to send and receive NL queries and answers, as well as the IDs of retrieved results. Based on these IDs, the cloud-based BIM rendered different 3D contextual scenes of the retrieved building objects in response to the user's query. Additionally, our web-based prototype also can be easily hosted by any cloud service provider (e.g., Google Cloud Platform) and ported to a web browser. This web-based setting reduces the requirements for high-performance workstations and allows for easy access from portable devices.

### 5.2 Discussion

Fig. 14 displays the developed VA prototype by implementing the BIM-GPT framework. Within the user interface, users input NL queries (shown in green boxes) and receive VA's NL answers (shown in white boxes) on the left in the chat boxes, while they visualize retrieved results and navigate the 3D model using the GUI of Autodesk Forge on the right. The screenshot demonstrates that the VA retrieved accurate information from the BIM, because the attribute values in the property window of the queried building component match the VA's response.

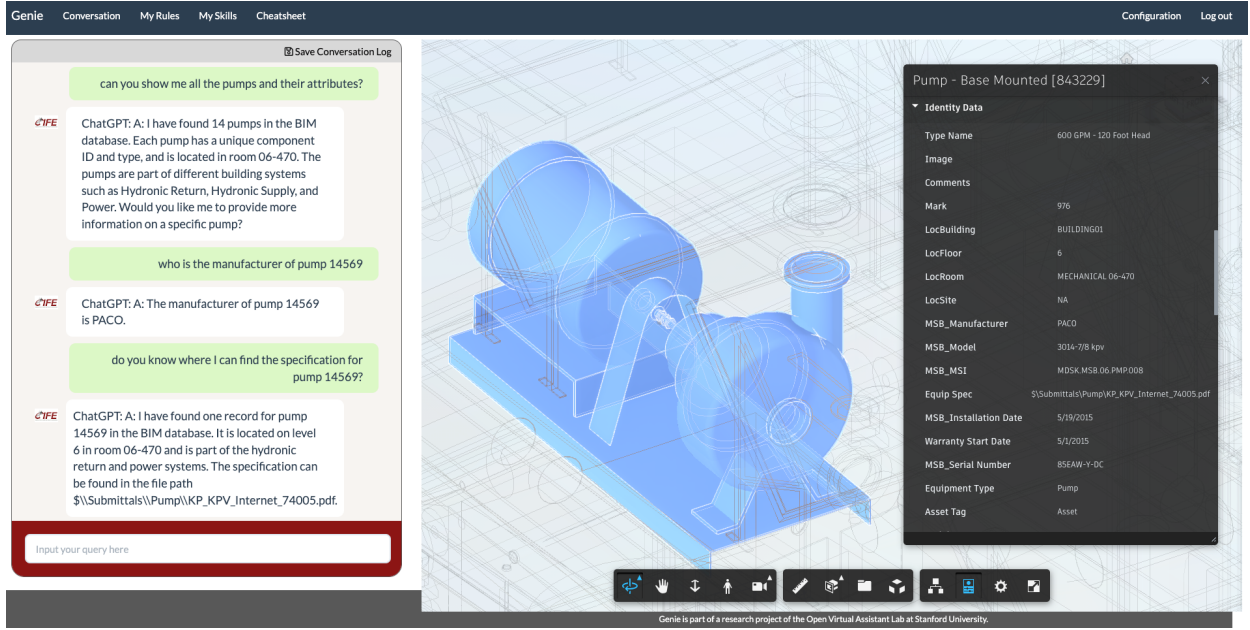

**Figure 14.** User Interface of VA Prototype: Chatbox (left); 3D Rendering (right)

In Fig. 15, a dialogue between the user and the VA is presented. Upon initializing the prototype, the BIM of the six-story hospital building was rendered. In this example, the user queried the building components by filtering on the level_number, component_type and component_id, and the corresponding retrieved results were visualized in the cloud-based BIM as shown in (a), (b), and (c), respectively. The summarized results, along with their 3D contexts, helped the user retrieve the required information and easily navigate the BIM. Finally, the user posed a general question about the building system attribute of the pump, and the VA's answer provided additional information, facilitating the IR process.

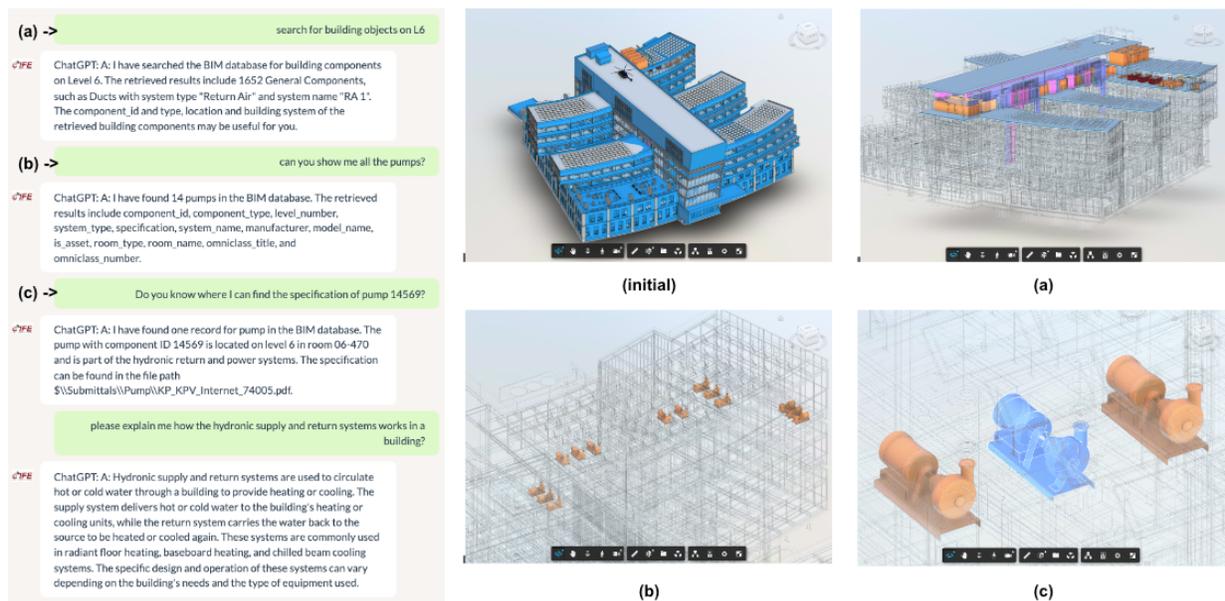

**Figure 15.** Example Conversations and Visualization of VA Prototype

The developed VA prototype validated the functionality of our BIM-GPT framework. The UI module enabled users to easily retrieve information through NL and navigate the cloud-based BIM. The NLP module accurately interpreted the user's queries, summarized retrieved information, and coherently answered BIM-related questions. The DM module successfully managed building information and provided retrieved results according to the user's queries. As such, our VA prototype reduced the BIM knowledge requirements for construction practitioners, supporting more efficient BIM IR.

In addition, the prototype development process demonstrated the significant benefits of our framework. By leveraging the capabilities of the GPT model, BIM-GPT substantially reduced the engineering effort required to handle NLP tasks. It eliminated the need for customizing syntactic and semantic analysis with traditional methods and the need for model training with ML methods, while accurately interpreting all NL queries during the testing process. For NLG, it eliminated the need to build predefined templates that only respond to queries within predefined syntactic patterns. For QA, it leveraged the GPT model as an external knowledge source, which required no additional effort for knowledge base construction. Furthermore, the VA's performance was easily improved by refining the prompts. This NL-based development enabled construction practitioners to participate in the rapid testing and prototyping processes, unlike existing approaches that heavily rely on technical experts to develop and maintain VAs.

## 5.3 Limitations

While we successfully implemented our novel framework, some limitations remain. Firstly, since this study was primarily focused on developing the framework, a quantitative evaluation of the framework was not included. In the future, we plan to conduct user studies to investigate the impact of the NL-based interface on the efficiency of BIM IR and evaluate the usability of the VA prototype. Secondly, the current framework required approximately 5 seconds to answer a BIM database search query, because of the need to make 5 ChatGPT API calls. Future research should reduce this latency while maintaining high accuracy in interpreting user queries. Thirdly, the current VA only supports single-turn conversations. This prevents follow-up conversation, which might be natural to users and useful for enhancing IR. Future research should investigate ways to support multi-turn conversations. Lastly, because GPT models might generate irrelevant or inaccurate information without proper prompting, future research should study how to improve the control of text generation, especially for NLG and QA tasks.

## 5. Conclusion

Efficiently retrieving information from BIMs is challenging because they contain a massive amount of multi-disciplinary data that are relevant across the building's lifecycle. Current retrieval strategies require users to have deep knowledge of BIM, which is a major impediment to utilizing the rich, up-to-date information of BIMs in practice. Although VA solutions have been explored to help address this accessibility issue, existing methods require extensive engineering efforts and data collection, which are generally not available or practical in the construction industry. This paper introduces BIM-GPT, a prompt-based virtual assistant (VA) framework that integrates BIM and GPT to enable practitioners to effectively query building information, receive summarized results with 3D visualization, and ask BIM-related questions using NL.

Our framework was evaluated on an augmented BIM IR dataset of 2065 NL queries. BIM-GPT demonstrated remarkable NLP capabilities and high data efficiency, achieving accuracy rates of 83.5% and 99.5% for classifying user intent and building object categories in NL queries with no data and 2% data incorporated in prompts, respectively. Additionally, BIM-GPT achieved accuracy rates as high as 97.8% and 88.9% for identifying building object parameters and recognizing parameter values in the NL queries, respectively, with similar prompt data levels. The results demonstrate that our prompt-based approach is effective and versatile in performing a wide range of NLP tasks.

To validate the functionality of the framework, we developed a VA prototype for a hospital building. Our implementation included a web-based interface with Autodesk Forge for BIM visualization, a prompt-based NLP module that connected with the ChatGPT API, and a MongoDB cloud-based database. The VA prototype was tested with 100 NL queries, and the results indicate that the VA effectively interacted with users through NL to support efficient BIM IR. This successful implementation underscores the potential of our BIM-GPT framework to support practical applications in the construction industry, improving the accessibility of BIMs for IR.

As such, our research contributes to the body of construction automation theory and practice. The BIM-GPT framework advances the theoretical understanding of how to leverage GPT models to automate BIM IR and how to design effective prompts for NLP tasks, such as text classification, information extraction, summarization, and question answering. By substantially reducing engineering and data requirements, the framework facilitates the development of VA for BIM to improve the speed, accuracy, and user experience of IR to support construction and facility operation activities. Its effectiveness and versatility also open opportunities for more advanced VAs and NL-based automation applications in the construction industry.

While the BIM-GPT framework shows promise, it is important to acknowledge its current limitations. The evaluation of the framework was conducted on a single dataset of NL queries, which may not represent all possible types of queries. The accuracy and effectiveness of the framework may vary depending on the size, diversity, and coverage of the dataset used for development and evaluation. Furthermore, this study mainly focused on the development of the framework, and as such, excludes user studies with construction practitioners. Therefore, further studies are needed to assess the impact of prompt-based VAs on the efficiency of BIM IR compared to existing retrieval strategies.

Nevertheless, this study represents an important step towards bridging the gap between NL-based IR and BIM through the development of the BIM-GPT framework. Future research should aim to improve the accuracy, efficiency, and generalizability of the framework. Additionally, BIM-GPT's functionality can be extended from IR to comprehensive management and operation of BIMs, allowing users to manipulate building information through NL. In the long-term, with continued development and refinement, the BIM-GPT framework has the potential to accelerate the digital transformation of the construction industry by supporting the development of prompt-based VAs for BIMs and NL-related automation applications.